\documentclass[10pt,twocolumn,letterpaper]{article}
\usepackage[pagenumbers]{cvpr}      %
\usepackage{cuted} %
\usepackage{currfile} %
\usepackage{caption} %

\usepackage{multirow}
\usepackage{colortbl}
\definecolor{cvprblue}{rgb}{0.21,0.49,0.74}
\usepackage[pagebackref,breaklinks,colorlinks,allcolors=cvprblue]{hyperref}
\def\confName{CVPR}
\def\confYear{2027}
\title{Selective Knowledge Control for Continual Learning of GUI Agents Over Application Streams}
\author{%
\normalsize Zirui Shang$^{1,2*}$, Xin Shu$^{2,3*}$, Yang Liu$^{2}$, Zhi Gao$^{1,2,4\dagger}$, Xinxiao Wu$^{1,4\dagger}$, Lifeng Fan$^{2}$\\[4pt]
\small $^1$Beijing Key Laboratory of Intelligent Information Technology,\\
\small School of Computer Science \& Technology, Beijing Institute of Technology\\
\small $^2$State Key Laboratory of General Artificial Intelligence, BIGAI\\
\small $^3$Wuhan University\\
\small $^4$Guangdong Laboratory of Machine Perception and Intelligent Computing,\\
\small Shenzhen MSU-BIT University
}
\hypersetup{pdftitle={Selective Knowledge Control for Continual Learning of GUI Agents Over Application Streams},
  pdfauthor={Zirui Shang, Xin Shu, Yang Liu, Zhi Gao, Xinxiao Wu, Lifeng Fan}}
\begin{document}
\maketitle
\begingroup
\renewcommand{\thefootnote}{\fnsymbol{footnote}}
\footnotetext[1]{These authors contributed equally.}
\footnotetext[2]{Corresponding authors: \texttt{gaozhibit@bit.edu.cn}, \texttt{wuxinxiao@bit.edu.cn}.}
\endgroup
\begin{abstract}
Continual learning is a crucial capability for Graphical User Interface (GUI) agents to adapt to evolving applications while retaining knowledge acquired from previous applications. Such application streams pose a challenging knowledge modeling problem: new applications often share underlying knowledge with past ones, yet also introduce specific knowledge that must not interfere with historical knowledge. In this paper, we propose activation-conditioned selective knowledge control, a lightweight method that achieves selective knowledge retention via neuron-level gradient manipulation. Our method maintains a compact historical knowledge state to protect highly activated MLP neurons that preserve previous knowledge. When a new application arrives, it performs real-time gradient surgery conditioned on forward activation. Concretely, the protected neurons are categorized into two types: unactivated neurons holding specific knowledge, whose gradients are truncated to prevent interference; and activated neurons holding shared knowledge, whose gradients are orthogonally projected to preserve stability while enabling adaptation. After each application stage, newly identified critical neurons are merged into the historical state for future learning. Empirical evaluations on multi-app sequential benchmark demonstrate that our method effectively mitigates catastrophic forgetting on prior applications while sustaining robust adaptation to new ones.
\end{abstract}
\section{Introduction}
\label{sec:intro}
\begin{figure}[t]
\centering
\includegraphics[width=\columnwidth]{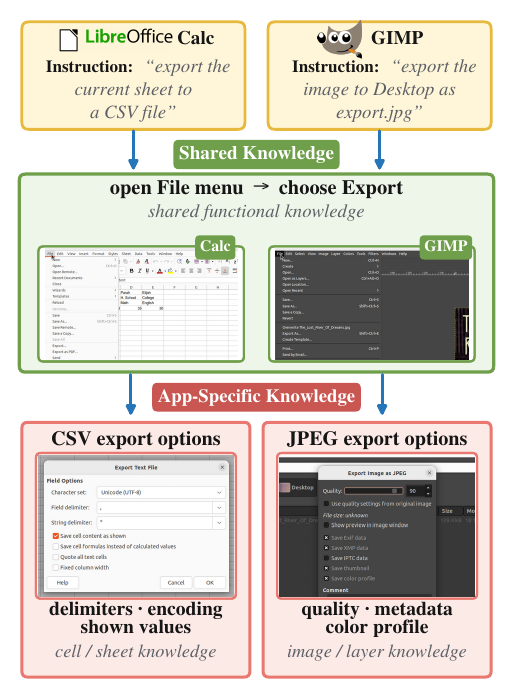}
\caption{Example of shared and application-specific knowledge across GUI tasks. Exporting a spreadsheet in LibreOffice Calc and an image in GIMP shares the procedure of opening the \texttt{File} menu, selecting \texttt{Export}, and interacting with an export dialog, while the subsequent steps are application-specific.
}
\label{fig:shared_specific}
\end{figure}

Graphical User Interface (GUI) agents powered by Multimodal Large Language Models (MLLMs) have emerged as a pivotal paradigm for automating complex interactions on desktops or mobiles. Recent systems and benchmarks have shown rapid progress in visual grounding, action prediction, and long-horizon computer use \citep{cheng2024seeclick,hong2024cogagent,zhou2023webarena,rawles2024androidworld,xie2025osworld,qin2025ui,shi2025guiknowledge,zhang2026tongui}. Yet practical GUI-agent deployment rarely conforms to a fixed training paradigm. New applications with changed layouts, user-specific workflows, and emerging functions continually introduce new interaction tasks. This creates a natural continual learning setting for GUI agents: the agents must adapt to new applications while retaining and reusing knowledge acquired from previous ones.
Continual learning of GUI agents over application streams raises a knowledge modeling problem. On one hand, a new application may rely on capabilities shared across diverse applications while also introducing application-specific behaviors that risk interfering with prior knowledge.
For instance, diverse applications share general GUI operations, such as locating visual widgets, reading screen text, opening menus, and filling fields. On the other hand, they demand distinct workflows, such as building spreadsheet formulas, scheduling calendar events, or handling application-specific export configurations, as shown in Fig.~\ref{fig:shared_specific}.
Consequently, continual GUI learning differs from traditional continual learning that only considers the trade-off between catastrophic forgetting and fast adaptation.
Continual GUI learning requires selectively safeguarding application-specific knowledge while keeping shared knowledge adaptable and reusable.
Existing continual-learning strategies, however, often treat historical knowledge uniformly when imposing preservation constraints, making it difficult to distinguish application-specific knowledge from shared knowledge. Parameter freezing and importance regularization methods~\citep{kirkpatrick2017overcoming} protect knowledge from previous applications, but they constrain protected components uniformly and may suppress updates to shared knowledge needed by a new application. Gradient surgery methods~\citep{yu2020gradient} further reduce destructive gradient interference by removing conflicting gradient components, but they do not distinguish reusable historical knowledge from knowledge that should be protected. Neuron-level protection methods are more fine-grained, but their fixed protected set still fails to distinguish protected and reusable historical knowledge \citep{serra2018hat,masana2021ternaryfeaturemasks}.
Prior studies have used neuron activation to identify neurons associated with factual knowledge, language-specific capabilities, and task-specific skills~\citep{dai2022knowledge,tang2024language,wang2022skill}. In GUI applications, we find that removing updates in neurons highly activated by only one application primarily reduces its performance, while removing updates in neurons highly activated by both applications reduces performance on both (Table~\ref{tab:application_neuron_ablation}). These findings motivate activation-based selective control of application-specific and shared historical knowledge.
In this paper, we propose activation-conditioned selective knowledge control, a lightweight method that moves beyond uniform knowledge management by executing selective, neuron-level gradient-based control over historical knowledge.
Given training data about a new application, the historical knowledge may play different roles: some knowledge is application-specific and should be preserved from unnecessary modification, while other knowledge is shared across applications and should remain adaptable and be further refined for the new application. The proposed method therefore treats knowledge control as a dynamic distinction between protected historical knowledge and reusable shared knowledge, rather than as a static preservation rule.
Specifically, our method maintains a compact historical knowledge state to protect highly activated MLP neurons with their historical update directions, which preserve previous knowledge.
During training on a new application, the proposed method performs activation-conditioned forward partitioning by registering a forward hook on MLP activations to collect neuron-level activation scores. During backward propagation, it performs neuron-level backward gradient surgery conditioned on the forward partition. For protected historical neurons that are not activated by the current update, their gradients are truncated to prevent unnecessary interference with application-specific historical knowledge. Meanwhile, for protected historical neurons that are activated by the current update, their gradients are projected onto the subspace orthogonal to cumulative historical update directions to preserve historical directions while allowing compatible adaptation. After each application training, a state update scheme is designed to merge newly identified important neurons into the historical knowledge state for future learning. The resulting procedure is lightweight, requiring no historical trajectory replay, old-application gradient computation, or full application-specific checkpoints.
Our contributions are as follows:
\begin{itemize}
    \item We propose activation-conditioned selective knowledge control for continual learning of GUI agents over application streams, which manages historical application knowledge selectively rather than uniformly by distinguishing application-specific knowledge from shared knowledge that should remain adaptable and reusable.
    \item We realize this method through neuron-level gradient operations conditioned on real-time activation: it truncates gradients of inactive protected neurons to prevent unnecessary interference and projects gradients of activated protected neurons orthogonal to cumulative historical update directions to allow compatible adaptation.
    \item Empirical evaluations on multi-app sequential benchmark demonstrate that the proposed method mitigates performance drops on previous applications while sustaining robust adaptation to new applications.
\end{itemize}
\section{Related Work}
\label{sec:related}
\subsection{GUI Agents}
GUI agents aim to understand screen observations, follow natural language instructions, and execute low-level actions such as clicking, typing, and navigation in digital environments. Existing systems can be roughly grouped into structured agents that rely on HTML, accessibility trees, APIs, or other metadata \citep{zhou2023webarena,deng2023mind2web,gur2023real,lai2024autowebglm}, visual agents that directly process screenshots with MLLMs \citep{cheng2024seeclick,you2024ferret,lu2024omniparser,xu2024aguvis,xie2026guide}, and hybrid agents that combine visual and structural signals \citep{he2024webvoyager,gou2024navigating,wu2024atlas,qin2025ui}. Concurrently, the training paradigm has evolved from supervised and data-centric fine-tuning on GUI demonstrations toward reinforcement fine-tuning from task outcomes or environment feedback \citep{hong2024cogagent,lin2024showui,chen2024guicourse,lu2024gui,ou2024synatra,xu2024agenttrek,putta2024agent,su2025learn,shi2026ira}. Meanwhile, benchmarks increasingly evaluate GUI agents across grounding, action prediction, multi-step task completion, and realistic computer-use workflows \citep{rawles2023androidinthewild,rawles2024androidworld,xie2025osworld,li2025screenspot,nayak2025ui,shi2025guiknowledge,zhang2026tongui}.
However, most training pipelines assume a fixed training data distribution under a one-time post-training paradigm, while practical agents must continually adapt to changing applications, layouts, workflows, and tool APIs while retaining previously learned interaction abilities.
\subsection{Continual Learning}
Continual learning studies how a model can learn from a sequence of tasks while maintaining useful prior knowledge. Classical approaches preserve previous knowledge through importance regularization \citep{kirkpatrick2017overcoming} or replay-based constraints \citep{lopezpaz2017gradient}. More closely related to our work are gradient-space methods that remove conflict gradients to avoid increasing losses on previous tasks \citep{lopezpaz2017gradient,farajtabar2020orthogonal,yu2020gradient}.
Beyond continual-learning algorithms, studies of MLP neurons provide evidence of functional specialization. Dai et al.~\citep{dai2022knowledge} investigate neurons associated with factual knowledge, Wang et al.~\citep{wang2022skill} identify task-specific skill neurons, and Tang et al.~\citep{tang2024language} identify language-specific neurons using activation statistics. These studies motivate our use of activation patterns for selective control of historical knowledge.
Recent work has started to study continual learning in GUI and computer-use agents, focusing on
domain and resolution shifts.
GUI-AiF \citep{liu2026continualgui}, GUI-AC \citep{lin2026guiac}, and ACuRL \citep{xue2026acurl}
improve continual adaptation by designing better rewards, optimization rules, or autonomous training data, but they do not directly control how new-application updates interact with historical GUI knowledge.
The closest work is Agent-Dice \citep{wu2026agentdice}, which addresses agent continual learning through post-hoc parameter fusion: it first trains on each application separately and then merges the resulting models.
In contrast, our method treats continual learning as a gradient-control problem for a single sequentially fine-tuned model. Rather than merging models, it
protects historical knowledge for the model while allowing shared functional neurons to adapt under an orthogonal constraint.
\section{Method}
\label{sec:method}
\begin{figure*}[t]
\centering
\includegraphics[width=0.83\textwidth]{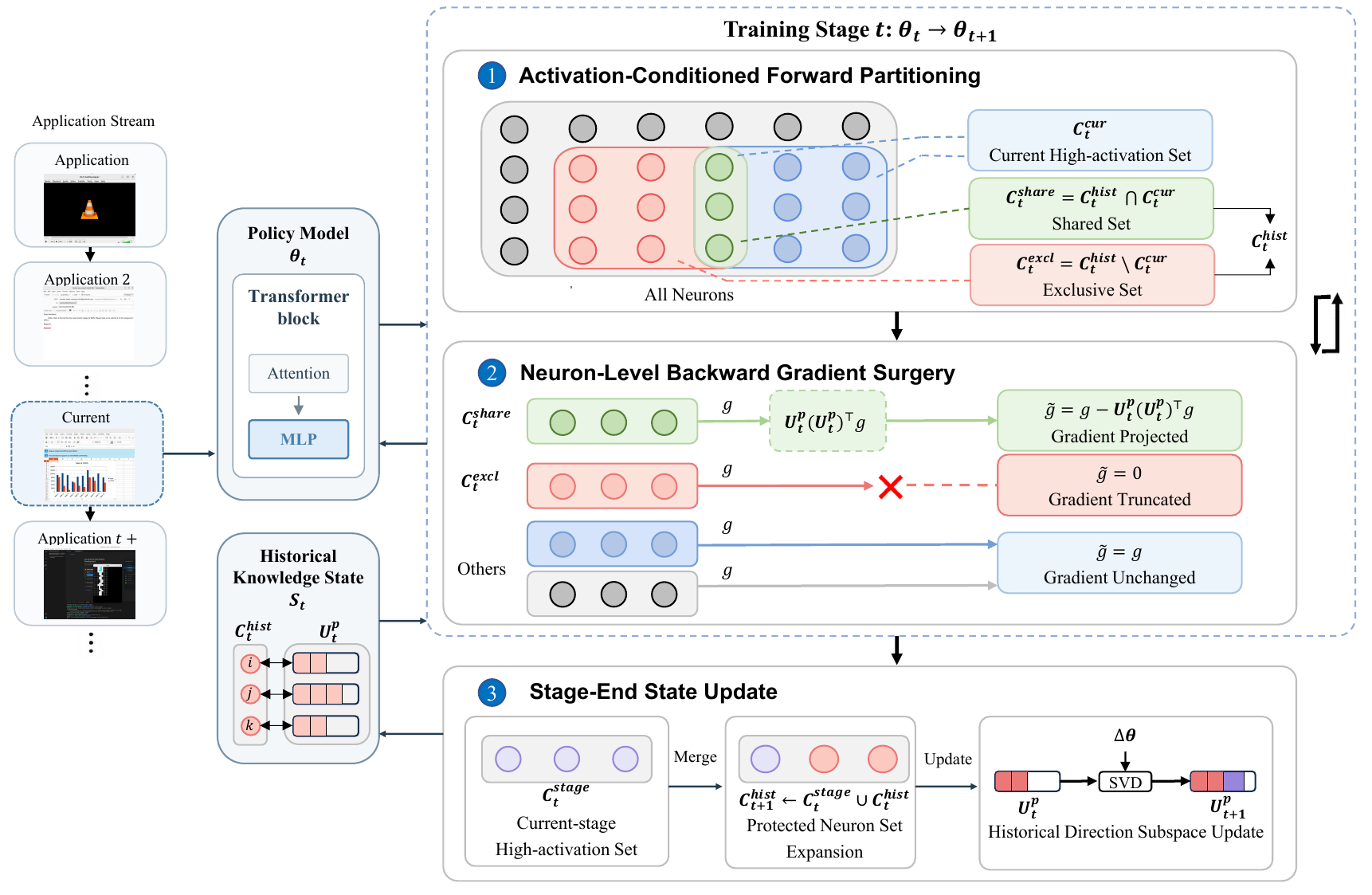}
\caption{Overview of the proposed activation-conditioned selective knowledge control framework.
Within each training stage, forward activations partition protected historical neurons into shared and application-specific subsets, and backward gradient surgery applies different update rules to these subsets. After the stage ends, the historical knowledge state is updated with newly identified important neurons and their historical direction subspaces.}
\label{fig:framework}
\end{figure*}
\subsection{Problem Setup and Formulation}
\label{sec:method_formulation}
We consider a GUI agent trained over an application stream with $N$ applications. Each stage contains all training tasks and trajectories from one application and is indexed by $t=1,\ldots,N$. At stage $t$, the agent fine-tunes a policy model $\pi_{\theta_t}$ initialized from the previous stage $t-1$. Let $\mathcal{L}_t(\theta)$ denote the training objective over the current application stage. Standard sequential fine-tuning updates the model on stage $t$ as follows,
\begin{equation}
    \theta_{t}^{k+1} = \theta_{t}^{k} - \eta g_t^k,
    \qquad
    g_t^k = \nabla_{\theta}\mathcal{L}_t(\theta_t^k),
\end{equation}
where $\theta_t^k$ denotes the model parameters at optimization step $k$ within stage $t$, and $g_t^k$ is the current gradient. The effect of update in stage $t$ on a historical objective $\mathcal{L}_i$ with $i<t$ can be approximated by a first-order Taylor expansion:
\begin{equation}
    \begin{aligned}
    \mathcal{L}_i(\theta_t^{k+1})
    &= \mathcal{L}_i(\theta_t^k - \eta g_t^k) \\
    &\approx \mathcal{L}_i(\theta_t^k) + \left\langle \nabla_{\theta}\mathcal{L}_i(\theta_t^k),  - \eta g_t^k \right\rangle \\
    &= \mathcal{L}_i(\theta_t^k) - \eta \left\langle \nabla_{\theta}\mathcal{L}_i(\theta_t^k), g_t^k \right\rangle.
    \end{aligned}
\end{equation}
Here, $\nabla_{\theta}\mathcal{L}_i(\theta_t^k)$ is the gradient of historical application $i$ evaluated at the current parameters $\theta_t^k$. This expression reveals the root cause of update interference: when
$\left\langle \nabla_{\theta}\mathcal{L}_i(\theta_t^k),\, g_t^k \right\rangle < 0$,
the resulting first-order change
$-\eta\left\langle \nabla_{\theta}\mathcal{L}_i(\theta_t^k),\, g_t^k \right\rangle$
is positive, indicating that the current update conflicts with previous knowledge and increases the historical loss. Continual learning therefore aims to prevent such negative alignment and reduce the resulting increase in historical loss.
In our method, we explicitly distinguish between shared and application-specific knowledge across GUI applications.
We argue that certain model neurons encode shared knowledge that should be continuously reinforced and updated along the application stream without increasing the loss on previous applications. Consequently, the update vector $g_t^k$ for these neurons must be orthogonal to the historical loss gradient $\nabla_{\theta}\mathcal{L}_i(\theta_t^k)$, ensuring that $\left\langle \nabla_{\theta}\mathcal{L}_i(\theta_t^k),\, g_t^k \right\rangle = 0$.
Conversely, for application-specific knowledge that requires no modification, the update $g_t^k$ on the corresponding neurons is set to zero to eliminate inter-application interference, naturally yielding $\left\langle \nabla_{\theta}\mathcal{L}_i(\theta_t^k),\, g_t^k \right\rangle = 0$.
Finally, for neurons that do not store historical knowledge, we directly apply the unmodified gradient $g_t^k$ for standard optimization.
In this case, our approach continuously refines shared knowledge while preserving previously acquired application-specific knowledge, and at the same time enabling efficient learning of specific knowledge for new applications.
Directly evaluating historical-task gradients at the current parameters would require access to historical training data. We instead use stage-wise parameter displacements as a compact proxy for historical update directions.
Specifically, we apply the method to MLP neurons in Transformer blocks. For clarity, let $d_{\mathrm{ff}}$ denote the MLP intermediate dimension, and let each neuron $j\in\{1,\ldots,d_{\mathrm{ff}}\}$ correspond to a parameter slice across the three MLP weight matrices: one column of $W^{\mathrm{down}}\in\mathbb{R}^{d_{\mathrm{model}}\times d_{\mathrm{ff}}}$, and one row of $W^{\mathrm{gate}},W^{\mathrm{up}}\in\mathbb{R}^{d_{\mathrm{ff}}\times d_{\mathrm{model}}}$. At the beginning of stage $t$, the method loads a historical knowledge state $\mathcal{S}_{t}$ updated after previous stages:
\begin{equation}
    \mathcal{S}_{t}
    =
    \left(C_{t}^{\mathrm{hist}}, U_{t}^{\mathrm{down}}, U_{t}^{\mathrm{gate}}, U_{t}^{\mathrm{up}}\right).
\end{equation}
Here, $C_{t}^{\mathrm{hist}} \subset \{1, \ldots, d_{\mathrm{ff}}\}$ tracks the set of neuron indices protected from past stages. For each projection $p \in \{\mathrm{down}, \mathrm{gate}, \mathrm{up}\}$, $U_{t}^p=\{U_{j,t}^p\}_{j\in C_{t}^{\mathrm{hist}}}$ maintains the corresponding historical update directions. Each element $U_{j,t}^p$ is an orthonormal basis matrix that spans the historical parameter-offset subspace for neuron $j$ in projection $p$.
The overall framework is shown in Fig.~\ref{fig:framework}, and the exact stage-end state update from $\mathcal{S}_{t-1}$ to $\mathcal{S}_t$ is detailed in Section~\ref{sec:method_state_update}.
\subsection{Activation-Conditioned Forward Partitioning}
\label{sec:method_partitioning}
During the forward phase of training on stage $t$, the proposed method registers forward hooks on the inputs of MLP down projections in Transformer blocks and computes a runtime activation score for each neuron:
\begin{equation}
    a_j = \frac{1}{BS}\sum_{b=1}^{B}\sum_{s=1}^{S}\left|h_{b,s,j}\right|,
\end{equation}
where $h_{b,s,j}$ denotes the activation of neuron $j$ for batch sample $b$ and sequence position $s$.
Rather than using a fixed numerical threshold, the proposed method selects the top $\lfloor \rho d_{\mathrm{ff}} \rfloor$ activated neurons to form the current high-activation set:
\begin{equation}
    C_t^{\mathrm{cur}}
    =
    \operatorname{Top}_{\rho}
    \left(\{a_j\}_{j=1}^{d_{\mathrm{ff}}}\right),
\end{equation}
where $\rho \in (0, 1]$ is the hyperparameter specifying the selection ratio.
The protected historical neurons are then partitioned according to whether they are reused by the current update:
\begin{equation}
    C_t^{\mathrm{share}} = C_{t}^{\mathrm{hist}}\cap C_t^{\mathrm{cur}},
    \qquad
    C_t^{\mathrm{excl}} = C_{t}^{\mathrm{hist}}\setminus C_t^{\mathrm{cur}}.
\end{equation}
This forward partitioning assigns neurons in $C_t^{\mathrm{excl}}$ to application-specific historical regions: they were important for previous application stages but are not actively used by the current update. In addition, the method assigns neurons in $C_t^{\mathrm{share}}$ to shared functional regions: they support historical application knowledge and are also activated by the current update, so they should remain trainable under a protection constraint.
\subsection{Neuron-Level Backward Gradient Surgery}
\label{sec:method_surgery}
During the back-propagation phase, the proposed method modifies the gradients of the MLP projections at the neuron level before parameter updates. Let $\nabla_{W_p}\mathcal{L}_t$ denote the full gradient of the current loss $\mathcal{L}_t$ with respect to weight matrix $W_p$ for $p \in \{\mathrm{down}, \mathrm{gate}, \mathrm{up}\}$. For each neuron $j$, its corresponding gradient slice is given by
\begin{equation}
    \begin{aligned}
    g_j^{\mathrm{down}}
    &=
    \left(\nabla_{W^{\mathrm{down}}}\mathcal{L}_t\right)[:,j],\\
    g_j^{\mathrm{gate}}
    &=
    \left(
    \left(\nabla_{W^{\mathrm{gate}}}\mathcal{L}_t\right)[j,:]
    \right)^\top,\\
    g_j^{\mathrm{up}}
    &=
    \left(
    \left(\nabla_{W^{\mathrm{up}}}\mathcal{L}_t\right)[j,:]
    \right)^\top.
    \end{aligned}
\end{equation}
For application-specific historical neurons $j\in C_t^{\mathrm{excl}}$, the proposed method blocks unnecessary current updates:
\begin{equation}
    \tilde{g}_j^p = 0,
    \qquad j\in C_t^{\mathrm{excl}}.
\end{equation}
This prevents the current update from modifying protected neurons that are not activated at the current stage.
For neurons $j \in C_t^{\mathrm{share}}$ with shared knowledge, we project the current gradient $g_j^p$ onto the orthogonal complement of the historical direction subspace spanned by $U_{j,t}^p$:
\begin{equation}
  \tilde{g}_j^p
  =
  g_j^p - U_{j,t}^p (U_{j,t}^p)^\top g_j^p,
  \qquad j\in C_t^{\mathrm{share}}.
\end{equation}
The remaining gradient is orthogonal to the historical update subspace and therefore preserves historical directions while still allowing the shared neuron to adapt during new-application training.
For unprotected neurons $j\notin C_{t}^{\mathrm{hist}}$, the proposed method leaves the gradient unchanged. The optimizer then updates the model using the modified gradients $\tilde{g}$.
\subsection{Stage-End State Update}
\label{sec:method_state_update}
After completing stage $t$, we perform a stage-end state update that refreshes both the protected-neuron set $C_t^{\mathrm{hist}}$ and the historical direction subspaces $U_t^p$ for $p \in \{\mathrm{down}, \mathrm{gate}, \mathrm{up}\}$.
\paragraph{Protected Neuron Set Expansion.} Unlike activation-conditioned forward partitioning, which uses runtime activations during training iterations, the stage-end update uses stage-level activation statistics collected over the completed application stage. Let $A_{j,t}$ denote the accumulated activation score of neuron $j$ during stage $t$. The current stage-level important-neuron set is selected as
\begin{equation}
    C_t^{\mathrm{stage}}
    =
    \operatorname{Top}_{\rho}
    \left(\{A_{j,t}\}_{j=1}^{d_{\mathrm{ff}}}\right).
\end{equation}
The historical protected set is then expanded by
\begin{equation}
    C_{t+1}^{\mathrm{hist}}
    =
    C_{t}^{\mathrm{hist}} \cup C_t^{\mathrm{stage}}.
\end{equation}
\paragraph{Historical Direction Subspace Update.} To update the historical direction subspaces, the stage-end update first evaluates the parameter displacement of each protected neuron $j \in C_{t+1}^{\mathrm{hist}}$ induced by stage $t$. For each projection $p \in \{\mathrm{down}, \mathrm{gate}, \mathrm{up}\}$, let $w_{j,t}^p$ and $w_{j,t+1}^p$ denote the parameter vector slice of neuron $j$ before and after training on stage $t$, respectively. The stage-induced parameter offset is then defined as $\Delta w_{j,t+1}^p = w_{j,t+1}^p - w_{j,t}^p$.
We then collect its offset vectors across all stages in which this neuron was selected as stage-important to construct a historical offset matrix:
\begin{equation}
    \mathbf{D}_{j,t+1}^p
    =
    \left[ \, \Delta w_{j,r}^p \, \right]_{r \in \mathcal{I}_{j,t}},
    \mathcal{I}_{j,t} = \{\, r \le t \mid j \in C_r^{\mathrm{stage}} \,\}.
\end{equation}
We then perform truncated Singular Value Decomposition on $\mathbf{D}_{j,t+1}^p$ and keep the top $r_{\mathrm{svd}}$ left singular directions:
\begin{equation}
    \mathbf{D}_{j,t+1}^p \approx U_{j,t+1}^p \Sigma_{j,t+1}^p (V_{j,t+1}^p)^\top.
\end{equation}
The truncated left singular matrix $U_{j,t+1}^p$ forms a compact orthonormal basis for the dominant historical parameter-displacement subspace of neuron $j$, with at most $r_{\mathrm{svd}}$ directions retained. It is stored into the updated collection $U_{t+1}^p=\{U_{j,t+1}^p\}_{j\in C_{t+1}^{\mathrm{hist}}}$.
The historical knowledge state is updated as
\begin{equation}
    \mathcal{S}_{t+1}
    =
    \left(C_{t+1}^{\mathrm{hist}}, U_{t+1}^{\mathrm{down}}, U_{t+1}^{\mathrm{gate}}, U_{t+1}^{\mathrm{up}}\right).
\end{equation}
This stage-end state update is performed between training stages, while online training only requires collecting current activations and modifying current gradients. Therefore, the proposed method does not replay historical trajectories, compute old-application gradients during new-application training, or maintain full application-specific checkpoints.
\begin{table*}[t]
\caption{Average success rates (\%) over three runs on all seen applications after each training stage. Parentheses show signed percentage-point differences between ours and the baseline. Column-best results are bolded, and ``--'' denotes unavailable results.}
\label{tab:prefix_average}
\centering
\setlength{\tabcolsep}{1.8pt}
\renewcommand{\arraystretch}{1.10}
\scriptsize
\begin{tabular}{@{}llccccccccc@{}}
\toprule
\textbf{Method}
& \textbf{Stage}
& \textbf{VLC}
& \textbf{Thunderbird}
& \textbf{Writer}
& \textbf{Impress}
& \textbf{Calc}
& \textbf{GIMP}
& \textbf{VS Code}
& \textbf{Chrome}
& \textbf{Overall} \\
\midrule
\multirow{8}{*}{Baseline}
& Stage 1 & 37.3 & -- & -- & -- & -- & -- & -- & -- & 37.3 \\
& Stage 2 & 33.3 & 42.2 & -- & -- & -- & -- & -- & -- & 37.5 \\
& Stage 3 & 29.4 & 33.3 & 47.6 & -- & -- & -- & -- & -- & 37.7 \\
& Stage 4 & 33.3 & 42.2 & 46.0 & 37.9 & -- & -- & -- & -- & 39.5 \\
& Stage 5 & 25.5 & 48.9 & 39.7 & 42.4 & 15.1 & -- & -- & -- & 31.9 \\
& Stage 6 & 23.5 & 53.3 & 33.3 & 45.5 & 13.5 & 57.7 & -- & -- & 35.5 \\
& Stage 7 & 21.6 & 42.2 & 39.7 & 30.3 & 9.2 & 60.3 & 49.3 & -- & 32.6 \\
& Stage 8 & 31.4 & 53.3 & 38.1 & 40.9 & 13.5 & 65.4 & 50.7 & 31.7 & 37.3 \\
\midrule
\multirow{8}{*}{Agent-Dice}
& Stage 1 & 37.3 & -- & -- & -- & -- & -- & -- & -- & 37.3 \\
& Stage 2 & 33.3 & 48.9 & -- & -- & -- & -- & -- & -- & 40.6 \\
& Stage 3 & 31.4 & 50.0 & 42.8 & -- & -- & -- & -- & -- & 41.2 \\
& Stage 4 & 24.1 & 53.3 & 41.3 & 45.1 & -- & -- & -- & -- & 41.9 \\
& Stage 5 & \textbf{38.0} & 46.7 & 47.6 & 37.7 & 10.3 & -- & -- & -- & 31.1 \\
& Stage 6 & 20.8 & 51.1 & 44.5 & 40.7 & 15.6 & 58.3 & -- & -- & 35.8 \\
& Stage 7 & 25.5 & 44.4 & 41.3 & 39.2 & 16.3 & 54.5 & 59.4 & -- & 37.7 \\
& Stage 8 & 35.3 & 48.9 & 49.2 & 44.6 & 18.4 & 60.3 & 56.5 & \textbf{41.5} & 41.7 \\
\midrule
\rowcolor{gray!12}
& Stage 1 & 37.3 & -- & -- & -- & -- & -- & -- & -- & 37.3 \\
\rowcolor{gray!12}
& Stage 2 & 37.3\,(+3.9) & 51.1\,(+8.9) & -- & -- & -- & -- & -- & -- & 43.8\,(+6.2) \\
\rowcolor{gray!12}
& Stage 3 & 35.3\,(+5.9) & \textbf{57.8}\,(+24.4) & \textbf{57.1}\,(+9.5) & -- & -- & -- & -- & -- & \textbf{50.3}\,(+12.6) \\
\rowcolor{gray!12}
& Stage 4 & 37.3\,(+3.9) & 55.6\,(+13.3) & 54.0\,(+7.9) & 47.3\,(+9.4) & -- & -- & -- & -- & 48.3\,(+8.8) \\
\rowcolor{gray!12}
& Stage 5 & 37.3\,(+11.8) & 55.6\,(+6.7) & 55.6\,(+15.9) & 39.4\,(-3.0) & \textbf{24.3}\,(+9.2) & -- & -- & -- & 38.3\,(+6.4) \\
\rowcolor{gray!12}
& Stage 6 & 29.4\,(+5.9) & 51.1\,(-2.2) & 47.6\,(+14.3) & \textbf{53.0}\,(+7.6) & 21.3\,(+7.8) & 62.8\,(+5.1) & -- & -- & 42.5\,(+7.1) \\
\rowcolor{gray!12}
& Stage 7 & 31.4\,(+9.8) & 44.4\,(+2.2) & 47.6\,(+7.9) & 41.2\,(+10.9) & 21.3\,(+12.1) & \textbf{70.5}\,(+10.3) & \textbf{65.2}\,(+15.9) & -- & 43.3\,(+10.6) \\
\rowcolor{gray!12}
\multirow{-8}{*}{Ours}
& Stage 8 & 35.3\,(+3.9) & 51.1\,(-2.2) & 47.6\,(+9.5) & 42.0\,(+1.1) & 21.4\,(+8.0) & 67.9\,(+2.6) & 63.8\,(+13.0) & 37.4\,(+5.7) & 42.7\,(+5.4) \\
\bottomrule
\end{tabular}
\end{table*}
\section{Experiments}
\label{sec:exp}
\subsection{Experimental Setup}
\paragraph{Benchmark and Application Stream.}
We evaluate the proposed method on OSWorld using UI-TARS-1.5-7B as the base model for GUI agents. To simulate continual learning over applications, we construct an 8-stage application stream consisting of \texttt{vlc}, \texttt{Thunderbird}, \texttt{libreoffice\_writer}, \texttt{libreoffice\_impress}, \texttt{LibreOffice Calc}, \texttt{gimp}, \texttt{VSCode}, and \texttt{chrome}. Each stage $t$ corresponds to an application in the stream.
After training on stage $t$, the agent is evaluated on all seen applications up to stage $t$. The updated model and the historical knowledge state are then carried forward to initialize stage $t+1$.
\paragraph{Implementation Details and Baseline.}
We build on the DART-GUI infrastructure~\citep{li2025dart} and train UI-TARS-1.5-7B on OSWorld using Group Relative Policy Optimization (GRPO)~\citep{shao2024deepseekmath}, with task-completion rewards from OSWorld. We keep the released training and evaluation pipeline unchanged except for our selective knowledge control module. For our method, we set the neuron selection ratio as $\rho=2\%$ and the truncated SVD rank as $r_{\mathrm{svd}}=4$. For comparison at each stage $t$, we construct a naive sequential fine-tuning baseline under the same infrastructure and optimization configuration, but without activation-conditioned selective knowledge control. We additionally compare with Agent-Dice~\citep{wu2026agentdice}, training each application-specific model independently from the same UI-TARS-1.5-7B checkpoint. At stage $t$, we fuse the parameter updates of the first $t$ application-specific models relative to this common base.
\paragraph{Evaluation Protocol.}
After each stage, we evaluate the agent three times on every application and report two groups of metrics:
(1) \textbf{Success rate.} OSWorld’s execution-based evaluators assign each trajectory a score in [0,1]. For each application, we report the average success rate over three evaluation runs. The Overall column reports the aggregate success rate over all seen applications, computed as the total number of successful evaluations divided by the total number of evaluation instances.
(2) \textbf{Retention--adaptation metrics.} We use retained performance (RP), current-application performance (CP), and retention--adaptation harmonic score (RAH) to separate historical retention from current adaptation. At a given stage, for each application learned in an earlier stage, let $s$ denote its current success rate and let $b$ denote its best success rate before this stage. RP quantifies historical knowledge retention by penalizing performance drops relative to prior peak scores as
$
\mathrm{RP}
=
1-\frac{\sum \max(b-s,0)}
{\sum b}.
$
CP is the success rate on the newly introduced application to measure current-stage adaptation performance, and RAH computes the harmonic mean of RP and CP to reflect the overall retention--adaptation balance.
\subsection{Main Results}
Table~\ref{tab:prefix_average} reports the average success rates over all applications observed up to each training stage. The proposed method improves the overall performance over naive sequential fine-tuning at every comparable stage from Stage 2 to Stage 8. The gains are especially clear in the middle and later stages, where the overall success rate increases by $12.6$, $7.1$, and $10.6$ percentage points at Stages 3, 6, and 7, respectively. These improvements indicate that activation-conditioned selective control helps the agent maintain stronger performance over the application stream instead of only adapting to the most recent application. Our method also outperforms Agent-Dice in overall success rate at every evaluated stage from Stage 2 to Stage 8. Across these stages, the average gain in overall success rate is $5.6$ percentage points. The corresponding standard deviations across three evaluation runs are reported in Appendix~\ref{app:evaluation_std}.
\begin{table}[t]
\caption{Retained performance (RP), current-application performance (CP), and retention--adaptation harmonic score (RAH) during Stages 2--8. Parentheses report the signed percentage-point difference (Ours$-$Baseline).}
\label{tab:retention_adaptation}
\centering
\setlength{\tabcolsep}{2.5pt}
\renewcommand{\arraystretch}{0.95}
\small
\begin{tabular}{@{}llccc@{}}
\toprule
\textbf{Stage} & \textbf{Method}
& \textbf{RP (\%)} & \textbf{CP (\%)}
& \textbf{RAH (\%)} \\
\midrule
\multirow{2}{*}{Stage 2}
& Baseline & 89.3 & 42.2 & 57.3 \\
& \cellcolor{gray!12}\textbf{Ours}
& \cellcolor{gray!12}100.0~(+10.7)
& \cellcolor{gray!12}51.1~(+8.9)
& \cellcolor{gray!12}67.6~(+10.3) \\
\midrule
\multirow{2}{*}{Stage 3}
& Baseline & 78.9 & 47.6 & 59.4 \\
& \cellcolor{gray!12}\textbf{Ours}
& \cellcolor{gray!12}97.7~(+18.9)
& \cellcolor{gray!12}57.1~(+9.5)
& \cellcolor{gray!12}72.1~(+12.7) \\
\midrule
\multirow{2}{*}{Stage 4}
& Baseline & 95.6 & 37.9 & 54.3 \\
& \cellcolor{gray!12}\textbf{Ours}
& \cellcolor{gray!12}96.5~(+0.9)
& \cellcolor{gray!12}47.3~(+9.4)
& \cellcolor{gray!12}63.5~(+9.2) \\
\midrule
\multirow{2}{*}{Stage 5}
& Baseline & 88.1 & 15.1 & 25.8 \\
& \cellcolor{gray!12}\textbf{Ours}
& \cellcolor{gray!12}94.2~(+6.1)
& \cellcolor{gray!12}24.3~(+9.2)
& \cellcolor{gray!12}38.6~(+12.9) \\
\midrule
\multirow{2}{*}{Stage 6}
& Baseline & 84.5 & 57.7 & 68.6 \\
& \cellcolor{gray!12}\textbf{Ours}
& \cellcolor{gray!12}87.9~(+3.4)
& \cellcolor{gray!12}62.8~(+5.1)
& \cellcolor{gray!12}73.3~(+4.7) \\
\midrule
\multirow{2}{*}{Stage 7}
& Baseline & 78.2 & 49.3 & 60.5 \\
& \cellcolor{gray!12}\textbf{Ours}
& \cellcolor{gray!12}85.1~(+6.8)
& \cellcolor{gray!12}65.2~(+15.9)
& \cellcolor{gray!12}73.8~(+13.3) \\
\midrule
\multirow{2}{*}{Stage 8}
& Baseline & 93.0 & 31.7 & 47.3 \\
& \cellcolor{gray!12}\textbf{Ours}
& \cellcolor{gray!12}90.1~(-2.9)
& \cellcolor{gray!12}37.4~(+5.7)
& \cellcolor{gray!12}52.9~(+5.6) \\
\midrule
\multirow{2}{*}{\textbf{Mean}}
& Baseline & 86.8 & 40.2 & 53.3 \\
& \cellcolor{gray!12}\textbf{Ours}
& \cellcolor{gray!12}\textbf{93.1~(+6.3)}
& \cellcolor{gray!12}\textbf{49.3~(+9.1)}
& \cellcolor{gray!12}\textbf{63.1~(+9.8)} \\
\bottomrule
\end{tabular}
\end{table}
Table~\ref{tab:retention_adaptation} further decomposes these gains into retention and current-application adaptation. The proposed method improves RP, CP, and RAH by $6.3$, $9.1$, and $9.8$ percentage points on average, respectively. This shows that the method reduces performance degradation on previously learned applications while still allowing effective adaptation to the newly introduced application.
To compare the proposed method with other continual-learning baselines under the same GRPO training configuration, Table~\ref{tab:app_replay_baseline} reports the stage-wise success rates and overall performance for the baseline, replay, EWC, LoRA, and the proposed method. During each training step on the current application, the replay baseline randomly selects one task from previous applications and incorporates one GRPO group of trajectories corresponding to the selected task into joint optimization. EWC~\citep{kirkpatrick2017overcoming} is implemented as importance regularization following the original formulation, and LoRA trains one adapter~\citep{hu2021lora} per application with an instruction-based router to select the active adapter at test time. These results show that replay helps preserve prior performance to some extent, but the proposed method remains more effective in balancing adaptation and retention.
\begin{table}[t]
\caption{Comparison under the same training configuration. The table reports stage-wise success rates and overall performance for the baseline, replay, EWC, LoRA, and the proposed method.}
\label{tab:app_replay_baseline}
\centering
\setlength{\tabcolsep}{1.2pt}
\renewcommand{\arraystretch}{1.08}
\footnotesize
\begin{tabular}{@{}llcccc@{}}
\toprule
\textbf{Method}
& \textbf{Stage}
& \textbf{VLC}
& \textbf{Thunderbird}
& \textbf{Writer}
& \textbf{Overall} \\
\midrule
\multirow{3}{*}{Baseline}
& Stage 1 & 37.3 & -- & -- & 37.3  \\
& Stage 2 & 33.3 & 42.2 & -- & 37.5 \\
& Stage 3 & 29.4 & 33.3 & 47.6 & 37.7 \\
\midrule
\multirow{3}{*}{Replay}
& Stage 1 & 37.3 & -- & -- & 37.3  \\
& Stage 2 & \textbf{39.5} & 40.5 & -- & 40.0 \\
& Stage 3 & 35.3 & 51.1 & 50.4 & 45.7 \\
\midrule
\multirow{3}{*}{EWC}
& Stage 1 & 37.3 & -- & -- & 37.3 \\
& Stage 2 & 29.2 & 53.5 & -- & 40.6 \\
& Stage 3 & 28.3 & 51.1 & \textbf{57.4} & 46.3 \\
\midrule
\multirow{1}{*}{LoRA}
& -- & 34.0 & 46.7 & 47.0 & 42.7 \\
\midrule
\rowcolor{gray!12}
& Stage 1 & 37.3 & -- & -- & 37.3  \\
\rowcolor{gray!12}
& Stage 2 & 37.3 & 51.1 & -- & 43.8  \\
\rowcolor{gray!12}
\multirow{-3}{*}{Ours}
& Stage 3 & 35.3 & \textbf{57.8} & 57.1 & \textbf{50.3}  \\
\bottomrule
\end{tabular}
\end{table}
\subsection{Ablation Studies}
\paragraph{Application-specific and shared neuron ablation.}
We examine the application-specific and shared neuron sets identified from activations on Thunderbird and Writer. Starting from the same post-Writer checkpoint, we remove the Writer-stage updates from Writer-specific neurons, the Thunderbird-stage updates from Thunderbird-specific neurons, and updates from both stages from shared neurons, leaving all other parameters unchanged. Table~\ref{tab:application_neuron_ablation} shows that removing updates from each application-specific region primarily reduces performance on the corresponding application, while removing updates from the shared region reduces performance on both. These results support the distinct application-specific and shared contributions of the neurons selected by activation.
\begin{table}[t]
\caption{Neuron region ablation for Thunderbird and Writer. Columns report task success rates (\%) on the two applications.}
\label{tab:application_neuron_ablation}
\centering
\small
\begin{tabular}{lcc}
\toprule
Ablated region & Thunderbird & Writer \\
\midrule
Unmodified & 57.8 & 57.1 \\
\midrule
Thunderbird-specific & 40.0 & 57.1 \\
Writer-specific & 53.3 & 38.1 \\
Shared & 46.7 & 52.4 \\
\bottomrule
\end{tabular}
\end{table}
\paragraph{Selective-control ablation.}
We compare four update-control settings to isolate the effect of activation-conditioned partitioning and neuron-level gradient surgery. The naive fine-tuning baseline leaves gradients unchanged for all protected neurons. The static projection variant removes activation-conditioned partitioning and projects gradients for all protected neurons. The static freezing variant truncates gradients for all protected neurons. Our method applies activation-aware selective control to protected neurons. Results are reported in Table~\ref{table:experiment_ablation}. The baseline achieves a mean success rate of 37.6\%. Static projection and static freezing improve the mean success rate to 41.5\% and 40.1\%, respectively. In contrast, our method reaches 47.1\% on average and also achieves the best performance on Stage 2 (43.8\%) and Stage 3 (50.3\%), showing that activation-aware selective control better balances adaptation and preservation.
\begin{table}[t]
\caption{Ablation studies on selective control with four update-control settings.}
\label{table:experiment_ablation}
\centering
\setlength{\tabcolsep}{2mm}
\renewcommand{\arraystretch}{1.12}
\small
\begin{tabular}{lccc}
\toprule
\textbf{Setting} & \textbf{Stage 2 (\%)} & \textbf{Stage 3 (\%)} & \textbf{Mean (\%)} \\
\midrule
Baseline           & 37.5 & 37.7 & 37.6 \\
Static Projection  & 42.7 & 40.3 & 41.5 \\
Static Freezing    & 41.7 & 38.4 & 40.1 \\
\midrule
\rowcolor{gray!12}
\textbf{Ours}      & \textbf{43.8} & \textbf{50.3} & \textbf{47.1} \\
\bottomrule
\end{tabular}
\end{table}
\begin{figure*}[t]
\centering
\includegraphics[width=0.8\textwidth]{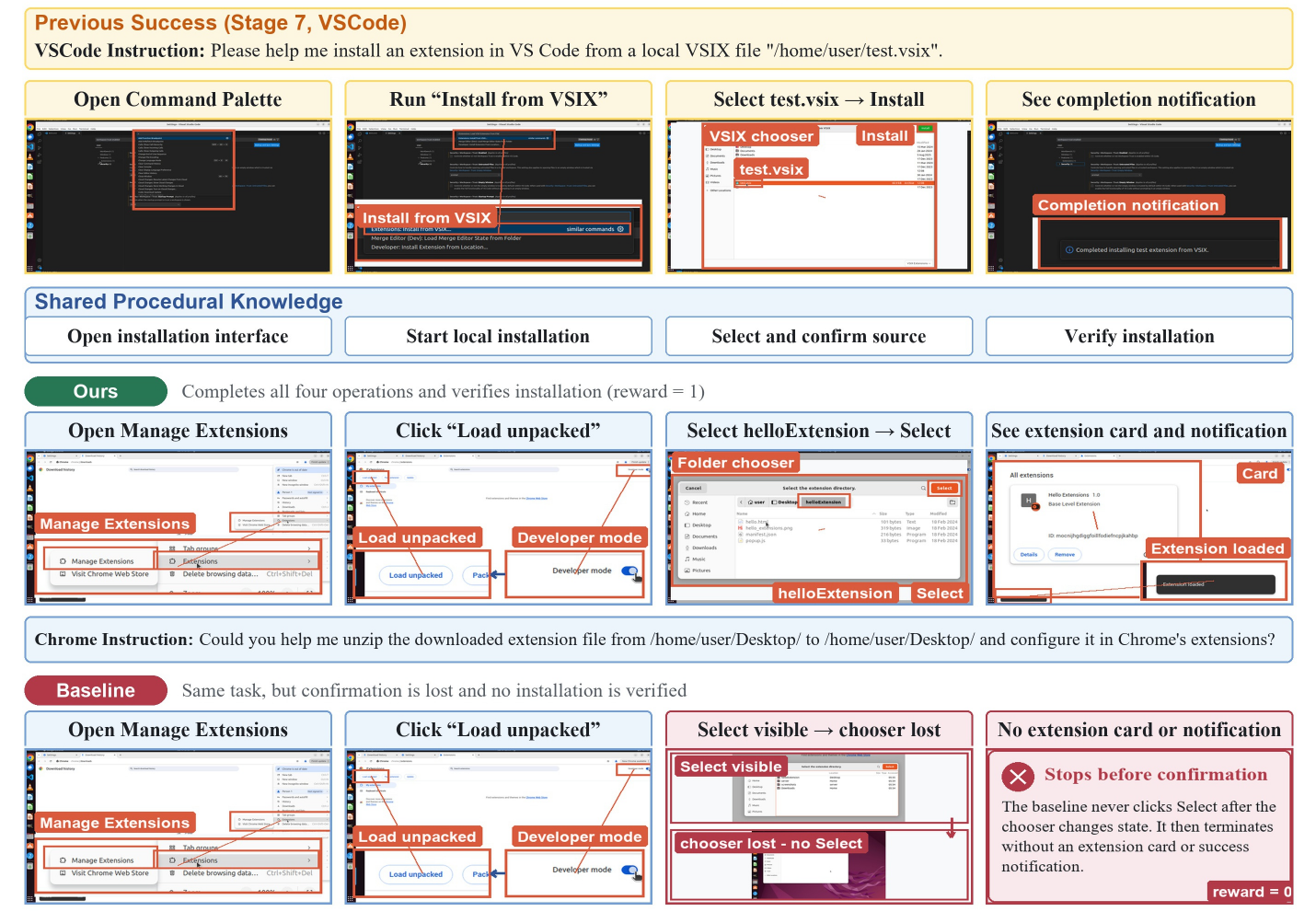}
\caption{Shared procedural knowledge reuse on a Chrome task.}
\label{fig:knowledge_case_study}
\end{figure*}
Additional analyses of hyperparameter sensitivity, application-stream order, and historical direction subspaces are provided in Appendices~\ref{app:hyperparameter}, \ref{app:training_order_comparison}, and~\ref{app:subspace_ablation}, respectively.
\subsection{Additional Analysis}
\paragraph{Training overhead.}
We measure the overhead introduced by activation-conditioned selective knowledge control during training. Table~\ref{table:experiment_throughput_single} compares the baseline and the proposed method under the same training configuration. Compared with the baseline, our method increases per-step training time from 4279.23s to 4324.00s, corresponding to a 1.0\% increase. Token throughput decreases by 3.3\%, and compute throughput decreases by 1.9\%. These results support the lightweight design of the proposed method, which adds little per-step time cost while preserving most training throughput under the same training framework.
\begin{table}[t]
\caption{Training overhead evaluation. We report per-step training time, token throughput (Tokens/s), and compute throughput (TFLOPs) under the same training configuration.}
\label{table:experiment_throughput_single}
\begin{center}
\small
\begin{tabular}{l | c c c}
\hline
\textbf{Method} & \textbf{Time (s/step)} & \textbf{Tokens/s} & \textbf{TFLOPs} \\
\hline
Baseline & 4279.23 & 400.07 & 289.79 \\
\rowcolor{gray!12}
\textbf{Ours} & 4324.00 & 386.73 & 284.30 \\
\hline
\end{tabular}
\end{center}
\end{table}
\paragraph{Case study on shared knowledge reuse.}
We provide a qualitative case showing how shared procedural knowledge acquired at Stage~7 is reused at Stage~8. As shown in Figure~\ref{fig:knowledge_case_study}, during the Stage~7 VSCode task, our method learns a general procedure from the successful \emph{Install from VSIX} workflow: open the installation interface, start local installation, select and confirm the source, and verify the installation. At Stage~8, this shared knowledge is applied to the Chrome task through application-specific controls: our method opens \emph{Manage Extensions}, clicks \emph{Load unpacked}, selects and confirms \emph{helloExtension}, and verifies the extension card and success notification, thereby completing the task. In contrast, the baseline loses the confirmation control after the chooser changes state and terminates without confirming the source or verifying the installation, resulting in task failure. This case illustrates the transfer of shared procedural knowledge learned from the Stage~7 VSCode workflow to the Stage~8 Chrome task.
Additional case studies and parameter-offset heatmap visualizations are provided in Appendix~\ref{app:additional_case_studies} and Appendix~\ref{app:parameter_offset_heatmap}, respectively.
\section{Conclusion}
This paper has studied continual learning for GUI agents over application streams, where GUI agents must adapt to new applications while retaining and reusing knowledge for previous ones.
By introducing activation-conditioned selective knowledge control, we move beyond uniform knowledge management through selective, neuron-level gradient-based control over historical knowledge, preserving application-specific knowledge while strengthening shared knowledge across applications.
Empirical evaluations on multi-app sequential benchmark demonstrate that the proposed method mitigates performance drops on previous applications while sustaining robust adaptation to new applications.

\section*{LLM Usage Statement}
OpenAI (ChatGPT / GPT) and Google (Gemini) assisted the preparation of this paper. The full manuscript, or major parts of it, was shared with these providers to obtain reviewer-style feedback. Their use on prose, related work and citations, experimental design and analysis, visual material, and the phrasing of ideas, hypotheses, claims, and conclusions was limited to editing or refining content written by the authors. Code and implementation include tool-generated portions that the authors reviewed, revised, and approved. These tools were not used for synthetic data, dataset creation, or annotation. The authors verified the outputs used in the submission and remain responsible for it.
{
    \small
    \bibliographystyle{ieeenat_fullname}
    \bibliography{references}
}
\clearpage
\maketitlesupplementary

\appendix
\section{Hyperparameter Analysis}
\label{app:hyperparameter}
We analyze the effect of two key hyperparameters: the selection ratio $\rho$ used for activation-conditioned partitioning and the truncated SVD rank $r_{\mathrm{svd}}$ used for historical direction storage. Results are shown in Tables~\ref{table:experiment_sensitivity_topk} and~\ref{table:experiment_sensitivity_svd}, respectively. The selection ratio table shows that $\rho=2\%$ performs best on the three observed applications, with an overall success rate of 50.3\%, while $\rho=1\%$ and $\rho=3\%$ reduce the overall success rate to 40.9\% and 43.7\%, respectively. The SVD-rank table shows that $r_{\mathrm{svd}}=4$ achieves the best overall success rate of 42.7\%, whereas $r_{\mathrm{svd}}=2$ and $r_{\mathrm{svd}}=8$ yield lower performance. These results indicate that the proposed method benefits from a moderate selection ratio and a moderate number of retained historical update directions.
\begin{table}[h]
\caption{Hyperparameter analysis of the selection ratio $\rho$.}
\label{table:experiment_sensitivity_topk}
\centering
\setlength{\tabcolsep}{2pt}
\small
\begin{tabular}{@{}lcccc@{}}
\toprule
\textbf{Selection Ratio} & \textbf{VLC} & \textbf{Thunderbird} & \textbf{Writer} & \textbf{Overall} \\
\midrule
$\rho=1\%$ & 27.5 & 51.1 & 44.4 & 40.9 \\
$\rho=3\%$ & 33.3 & 48.9 & 48.3 & 43.7 \\
\midrule
\rowcolor{gray!12}
\textbf{Ours ($\rho=2\%$)} & \textbf{35.3} & \textbf{57.8} & \textbf{57.1} & \textbf{50.3} \\
\bottomrule
\end{tabular}
\end{table}
\begin{table}[t]
\caption{Hyperparameter analysis of the truncated SVD rank $r_{\mathrm{svd}}$, reporting the overall success rate (\%) at Stage 8 under different numbers of retained historical update directions.}
\label{table:experiment_sensitivity_svd}
\centering
\small
\begin{tabular}{lccc}
\toprule
\textbf{SVD Rank}
& $r_{\mathrm{svd}}=2$
& \cellcolor{gray!12}\textbf{$r_{\mathrm{svd}}=4$}
& $r_{\mathrm{svd}}=8$ \\
\midrule
\textbf{Overall Success Rate (\%)}
& 38.5
& \cellcolor{gray!12}\textbf{42.7}
& 38.7 \\
\bottomrule
\end{tabular}
\end{table}
\section{Training Order Comparison}
\label{app:training_order_comparison}
We evaluate whether the proposed method remains stable under the shuffled application-stream order \texttt{VLC} $\rightarrow$ \texttt{VS Code} $\rightarrow$ \texttt{GIMP} compared with the baseline. Table~\ref{tab:app_training_order} shows that the proposed method still outperforms the baseline across stages under the shuffled order on overall performance, indicating that the advantage of selective control is robust to changes in application-stream order.
\begin{table}[h]
\caption{Application-stream order ablations. The table reports stage-wise success rates and overall performance of the baseline and the proposed method under a shuffled application-stream order.}
\label{tab:app_training_order}
\centering
\setlength{\tabcolsep}{1.2pt}
\renewcommand{\arraystretch}{1.08}
\footnotesize
\begin{tabular}{@{}llcccc@{}}
\toprule
\textbf{Method}
& \textbf{Stage}
& \textbf{VLC}
& \textbf{VS Code}
& \textbf{GIMP}
& \textbf{Overall} \\
\midrule
\multirow{3}{*}{Baseline}
& Stage 1 & 37.3 & -- & -- & 37.3 \\
& Stage 2 & 39.2 & 65.2 & -- & 54.2 \\
& Stage 3 & 41.2 & 59.4 & 66.2 & 57.4 \\
\cmidrule(l){1-6}
\multirow{3}{*}{Ours}
& Stage 1 & 37.3 & -- & -- & 37.3 \\
& Stage 2 & 39.2 & 69.6 & -- & 56.7 \\
& Stage 3 & 37.3 & 63.8 & 75.6 & 61.6 \\
\bottomrule
\end{tabular}
\end{table}
\section{Historical Subspace Ablation}
\label{app:subspace_ablation}
We ablate the historical directions used for gradient projection during Writer training after VLC and Thunderbird. Random replaces the historical directions with random directions while preserving their rank and norms. Shuffled reassigns historical directions among protected neurons within the same layer. No projection disables projection for shared neurons while retaining gradient truncation for historical exclusive neurons. Table~\ref{tab:subspace_ablation} shows that the proposed method achieves the highest overall and current-application performance, while exceeding the alternatives on the historical applications. Random and Shuffled outperform No projection in overall performance, but remain below the proposed method. These results support the importance of historical update directions and their correspondence to individual neurons for balancing historical knowledge retention and current-application adaptation.
\begin{table}[t]
\caption{Historical subspace ablation during Writer training after VLC and Thunderbird. Results are reported on all three seen applications.}
\label{tab:subspace_ablation}
\centering
\small
\setlength{\tabcolsep}{3pt}
\begin{tabular}{@{}lcccc@{}}
\toprule
Direction setting & VLC & Thunderbird & Writer & Overall \\
\midrule
Random & 27.9 & 57.8 & 52.4 & 46.1 \\
Shuffled & 26.8 & 55.6 & 54.0 & 45.7 \\
No projection & 22.2 & 44.4 & 54.0 & 41.1\\
\midrule
Ours & \textbf{35.3} & \textbf{57.8} & \textbf{57.1} & \textbf{50.3} \\
\bottomrule
\end{tabular}
\end{table}
\section{Additional Case Studies}
\label{app:additional_case_studies}
\begin{figure*}[t]
\centering
\includegraphics[width=\textwidth]{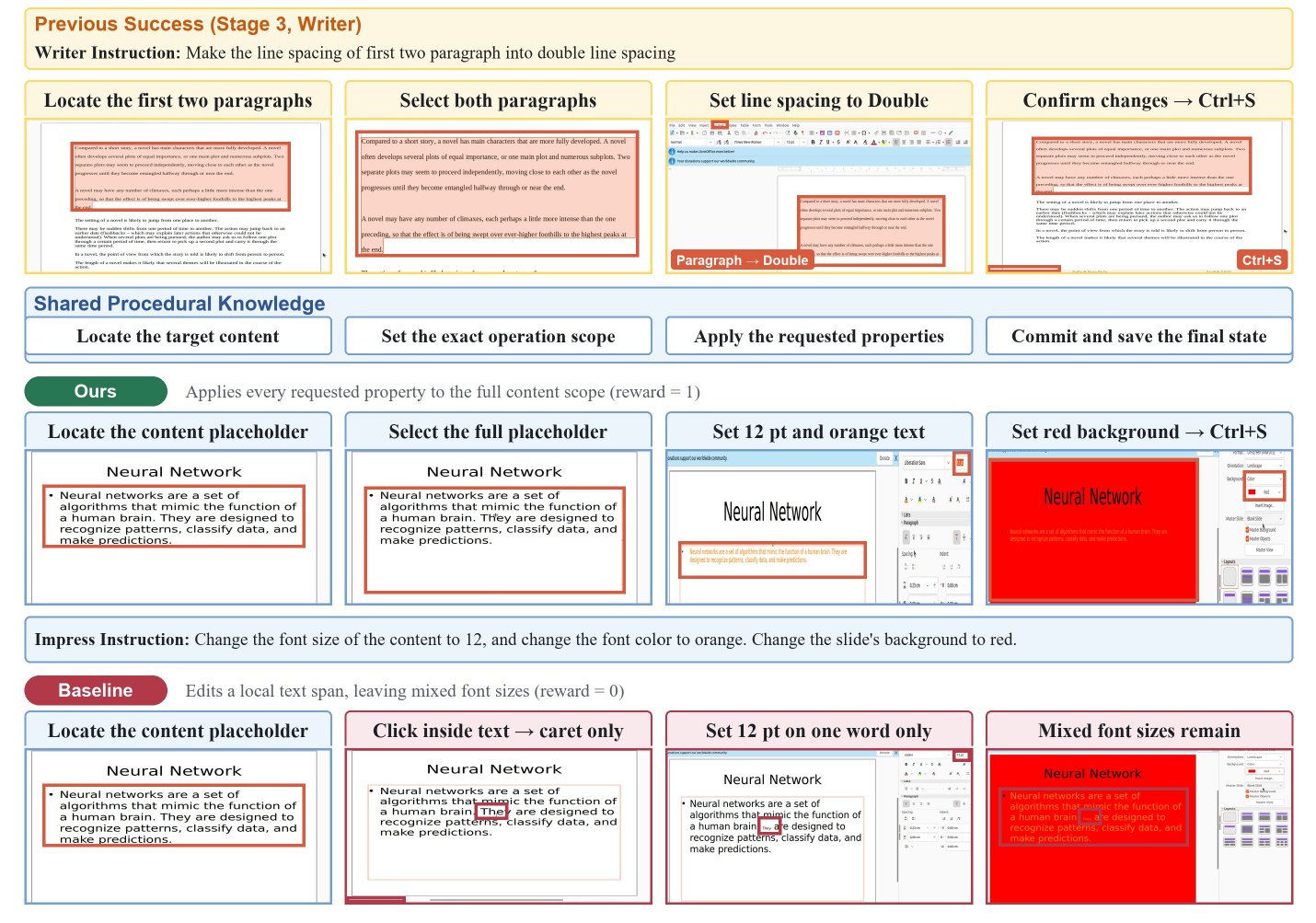}
\caption{Shared scope-aware formatting knowledge transferred from Stage~3 Writer to Stage~4 Impress.}
\label{fig:appendix_impress_formatting}
\end{figure*}

\begin{figure*}[t]
\centering
\includegraphics[width=\textwidth]{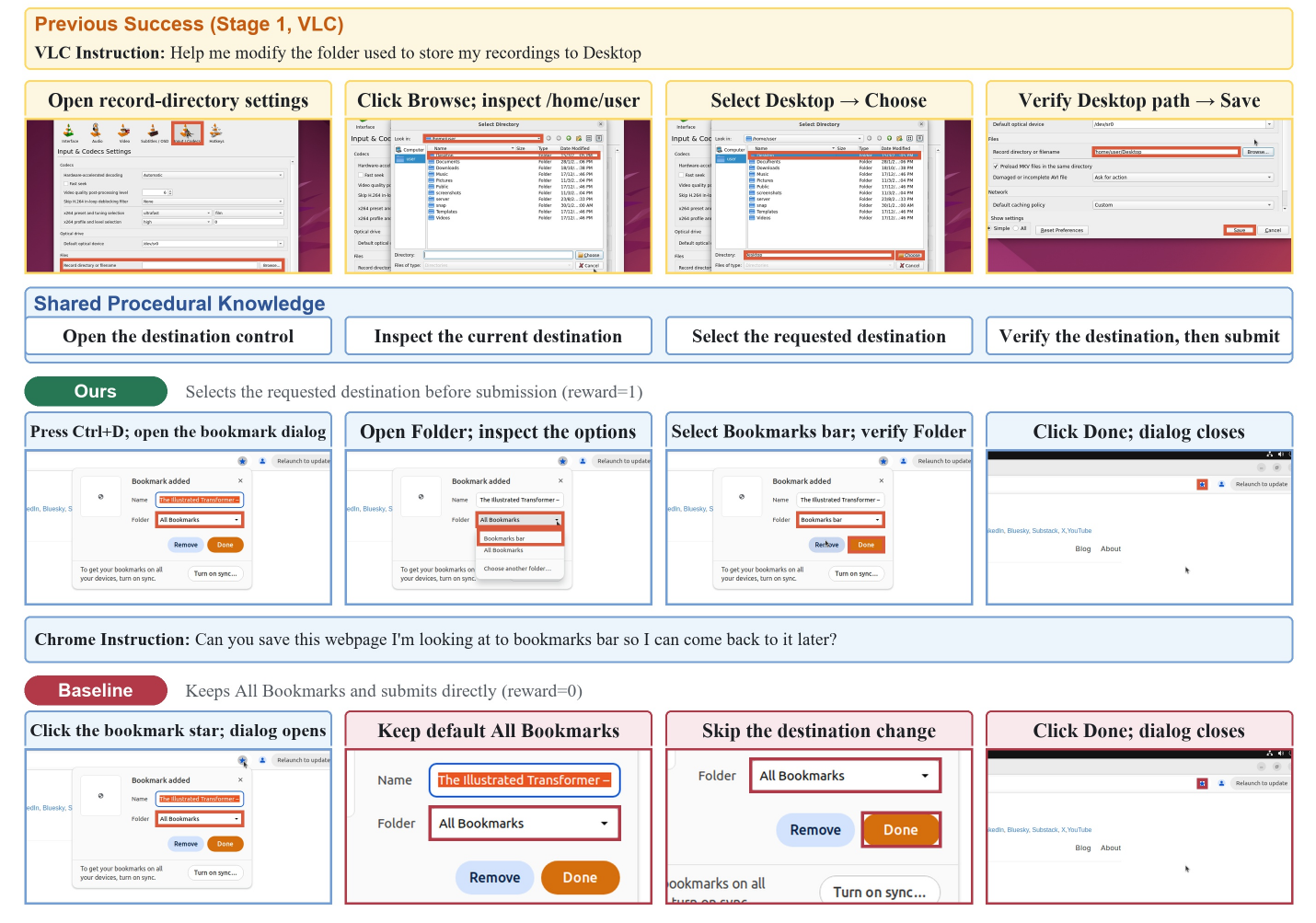}
\caption{Shared destination-aware procedural knowledge transferred from Stage~1 VLC to Stage~8 Chrome.}
\label{fig:appendix_chrome_bookmarks}
\end{figure*}

We provide two additional qualitative cases to further illustrate how procedural knowledge learned from earlier applications is reused in later applications with different interfaces and controls.
\paragraph{Scope-aware formatting from Writer to Impress.}
Figure~\ref{fig:appendix_impress_formatting} presents knowledge transfer from the Stage~3 Writer task to the Stage~4 Impress task. In Writer, the successful trajectory locates the first two paragraphs, selects the complete target scope, applies double line spacing, and saves the result. Our method transfers this scope-aware formatting procedure to Impress by locating the content placeholder, selecting the full placeholder, applying the requested 12-point orange text, setting the slide background to red, and saving the final state. The baseline locates the correct placeholder but places the caret inside the text instead of selecting the full content. Consequently, the font-size change affects only a local span and mixed font sizes remain, resulting in task failure. This case shows that reusable procedural knowledge includes determining the correct operation scope before applying the requested properties.
\paragraph{Destination-aware completion from VLC to Chrome.}
Figure~\ref{fig:appendix_chrome_bookmarks} provides a qualitative example of destination-aware procedural knowledge reused across applications. In the Stage~1 VLC workflow, our method selects the requested \emph{Desktop} directory, verifies that the displayed path has been updated, and only then saves the setting. This workflow establishes a general completion criterion: the task-specified destination should be reflected in the visible GUI state before submission. At Stage~8, both our method and the baseline open the same Chrome bookmark dialog. Our method checks the folder setting, changes the default \emph{All Bookmarks} to the requested \emph{Bookmarks bar}, verifies the updated value, and then clicks \emph{Done}. The baseline instead clicks \emph{Done} without changing or explicitly verifying the destination and consequently fails the task. The two trajectories therefore differ in whether the destination-specific goal state is verified before termination, which is consistent with the reuse of the procedure learned from VLC.
\section{Parameter-Offset Heatmap}
\label{app:parameter_offset_heatmap}
To examine how the proposed method changes the optimization trajectory, we compare stage-wise MLP parameter offsets with those of the baseline across all 28 decoder layers. For method
$m\in\{\mathrm{ours},\mathrm{base}\}$ and projection $p\in\{\mathrm{down},\mathrm{gate},\mathrm{up}\}$, the stage-induced parameter offset of decoder layer $\ell$ is defined by
\begin{equation*}
\begin{aligned}
\Delta W_{\ell,t}^{m,p}
&=
\operatorname{vec}(W_{\ell,t}^{m,p})
-
\operatorname{vec}(W_{\ell,t-1}^{m,p}),\\
\Delta_{\ell,t}^{m}
&=
\operatorname{concat}\!\left(
\Delta W_{\ell,t}^{m,\mathrm{down}},
\Delta W_{\ell,t}^{m,\mathrm{gate}},
\Delta W_{\ell,t}^{m,\mathrm{up}}
\right).
\end{aligned}
\end{equation*}
We characterize the directional and magnitude differences between the two offsets using
\begin{equation*}
\begin{aligned}
 D_{\ell,t}
&=
 \cos\!\left(
\Delta_{\ell,t}^{\mathrm{ours}},
\Delta_{\ell,t}^{\mathrm{base}}
\right),\\
R_{\ell,t}
&=
\frac{
 \left\|\Delta_{\ell,t}^{\mathrm{ours}}\right\|_2
}{
 \left\|\Delta_{\ell,t}^{\mathrm{base}}\right\|_2
 }.
\end{aligned}
\end{equation*}
Here, $D_{\ell,t}$ measures update direction similarity and $R_{\ell,t}$ compares update magnitudes. We visualize $\log_2 R_{\ell,t}$ so that zero denotes equal magnitudes, while positive and negative values indicate larger and smaller offsets under our method, respectively.
\begin{figure*}[t]
\centering
\includegraphics[width=\textwidth]{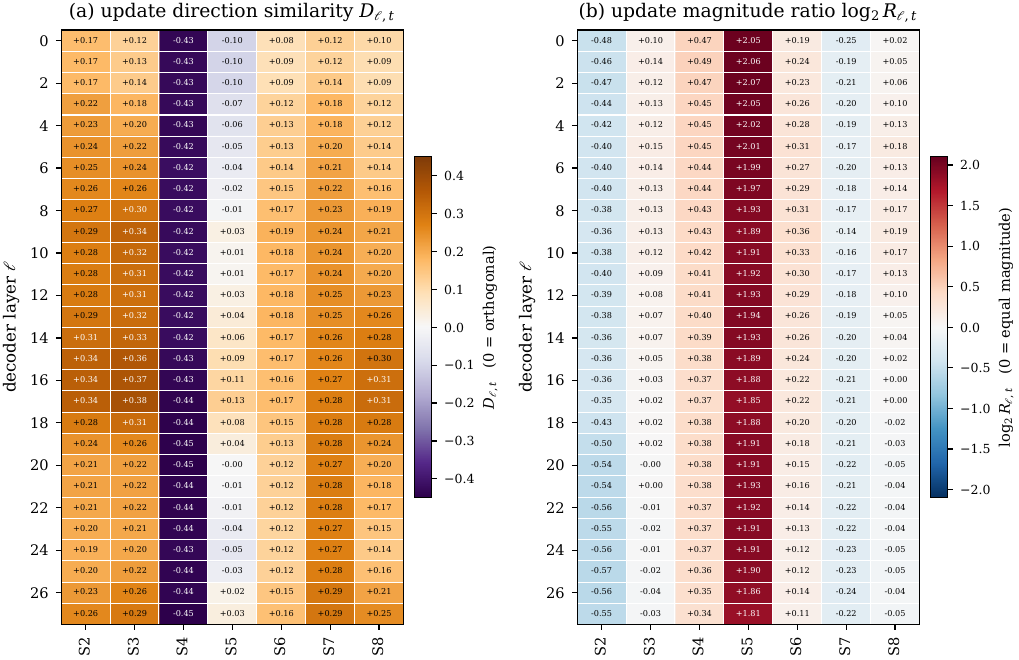}
\caption{Layer-wise comparison of stage-wise MLP parameter offsets between ours and the baseline. Rows correspond to the 28 decoder layers, indexed from 0 to 27.
(a) Direction similarity $D_{\ell,t}$, where zero denotes orthogonal directions.
(b) Log displacement ratio $\log_2 R_{\ell,t}$, where zero denotes equal update magnitudes.
S2--S8 correspond to Thunderbird, Writer, Impress, Calc, GIMP, VS Code, and Chrome, respectively.}
\label{fig:parameter_offset_heatmap}
\end{figure*}
Figure~\ref{fig:parameter_offset_heatmap}(a) shows that the proposed method changes the layer-wise update directions across the decoder stack. Over all $28\times7=196$ layer--stage pairs, the mean direction similarity is low, and the Stage~4 offsets consistently point in an opposing direction to those of the baseline, whereas the Stage~5 offsets are approximately orthogonal across most layers. Figure~\ref{fig:parameter_offset_heatmap}(b) further shows that this effect does not arise from uniformly smaller updates. The median displacement ratio is $1.088$, and only $37\%$ of the layer--stage pairs satisfy $R_{\ell,t}<1$. At Stage~5, for example, $\log_2 R_{\ell,t}$ ranges from $1.81$ to $2.07$, corresponding to offsets approximately $3.5$--$4.2$ times as large as those of the baseline. Overall, the proposed method mainly redirects layer-wise updates rather than merely suppressing their magnitude.
\section{Standard Deviations of Main Results}
\label{app:evaluation_std}
Table~\ref{tab:evaluation_std} reports the sample standard deviations of application-level and overall scores across three evaluation runs, complementing the mean results in Table~\ref{tab:prefix_average}
\begin{table*}[t]
\caption{Standard deviations across three evaluation runs corresponding to the mean results in Table~\ref{tab:prefix_average}. Values are in percentage points; ``--'' denotes applications not yet introduced.}
\label{tab:evaluation_std}
\centering
\small
\setlength{\tabcolsep}{3pt}
\begin{tabular}{@{}llccccccccc@{}}
\toprule
Method & Stage & VLC & Thunderbird & Writer & Impress & Calc & GIMP & VS Code & Chrome & Overall \\
\midrule
\multirow{8}{*}{Baseline}
 & Stage 1 & 6.79 & -- & -- & -- & -- & -- & -- & -- & 6.79 \\
 & Stage 2 & 8.99 & 10.18 & -- & -- & -- & -- & -- & -- & 8.27 \\
 & Stage 3 & 5.88 & 0.00 & 4.76 & -- & -- & -- & -- & -- & 3.77 \\
 & Stage 4 & 3.40 & 13.88 & 2.75 & 1.31 & -- & -- & -- & -- & 1.19 \\
 & Stage 5 & 3.40 & 10.18 & 2.75 & 6.56 & 2.31 & -- & -- & -- & 2.01 \\
 & Stage 6 & 0.00 & 6.67 & 4.76 & 2.27 & 3.25 & 5.88 & -- & -- & 2.45 \\
 & Stage 7 & 0.00 & 6.67 & 5.50 & 1.31 & 2.13 & 3.85 & 7.53 & -- & 0.30 \\
 & Stage 8 & 3.40 & 6.67 & 5.50 & 0.00 & 2.46 & 0.00 & 4.35 & 4.88 & 0.66 \\
\midrule
\multirow{8}{*}{Agent-Dice}
 & Stage 1 & 6.79 & -- & -- & -- & -- & -- & -- & -- & 6.79 \\
 & Stage 2 &   0.96 &   11.55 & -- & -- & -- & -- & -- & -- &  4.99 \\
 & Stage 3 &   6.79 &   3.85 &   0.00 & -- & -- & -- & -- & -- &  3.15 \\
 & Stage 4 &   8.98 &   7.70 &   5.02 &   5.03 & -- & -- & -- & -- &  4.16 \\
 & Stage 5 &   8.63 &   0.00 &   7.53 &   1.11 &   2.68 & -- & -- & -- &  2.08 \\
 & Stage 6 &   2.97 &   7.70 &   2.51 &   4.26 &   1.23 &   6.75 & -- & -- &  1.32 \\
 & Stage 7 &   10.38 &   10.18 &   5.02 &   1.35 &   1.88 &   4.00 &   2.51 & -- &  1.21 \\
 & Stage 8 &   8.99 &   0.00 &   0.00 &   1.84 &   3.25 &   2.22 &   13.10 &   6.97 &  2.05 \\
\midrule
\multirow{8}{*}{Ours}
 & Stage 1 & 6.79 & -- & -- & -- & -- & -- & -- & -- & 6.79 \\
 & Stage 2 & 5.88 & 7.70 & -- & -- & -- & -- & -- & -- & 6.51 \\
 & Stage 3 & 5.88 & 7.70 & 4.76 & -- & -- & -- & -- & -- & 2.88 \\
 & Stage 4 & 3.40 & 7.70 & 5.50 & 0.70 & -- & -- & -- & -- & 2.67 \\
 & Stage 5 & 3.40 & 10.18 & 9.91 & 1.31 & 1.55 & -- & -- & -- & 0.58 \\
 & Stage 6 & 5.88 & 3.85 & 4.76 & 2.62 & 2.13 & 5.88 & -- & -- & 2.07 \\
 & Stage 7 & 3.40 & 3.85 & 4.76 & 2.79 & 2.13 & 15.54 & 4.35 & -- & 2.81 \\
 & Stage 8 & 6.79 & 3.85 & 7.27 & 3.41 & 2.14 & 7.69 & 6.64 & 3.73 & 1.51 \\
\bottomrule
\end{tabular}
\end{table*}

\end{document}